\documentclass{article}

\usepackage{arxiv}
\usepackage{natbib}
\usepackage[utf8]{inputenc}
\usepackage{hyperref}
\usepackage{booktabs}
\usepackage{amsfonts}
\usepackage{nicefrac}
\usepackage{microtype}
\usepackage{doi}
\usepackage{graphicx}
\usepackage{caption}
\usepackage{latexsym}
\usepackage{amssymb}
\usepackage{amsmath}
\usepackage{amsthm}
\usepackage{booktabs}
\usepackage{enumitem}
\usepackage{graphicx}
\usepackage{color}
\usepackage[dvipsnames]{xcolor}
\usepackage{subfigure}
\usepackage{cleveref}
\usepackage{multirow}
\usepackage{adjustbox}
\usepackage{nicefrac}
\usepackage{algorithm}
\usepackage{algorithmic}
\usepackage{url}

\newcommand{\PreserveBackslash}[1]{\let\temp=\\#1\let\\=\temp}
\DeclareMathOperator*{\argmax}{arg\,max}
\DeclareMathOperator*{\argmin}{arg\,min}
\newcommand{\ci}{\perp\!\!\!\perp}

\title{Robust Non-Linear Correlations via Polynomial Regression}

\author{
    \href{https://orcid.org/0000-0002-9120-6949}{Luca Giuliani} \\
    Department of Computer Science and Engineering \\
    University of Bologna \\
    Viale del Risorgimento 2, Bologna (BO), Italy \\
    \texttt{luca.giuliani13@unibo.it} \\
    \And
    \href{https://orcid.org/0000-0003-4709-8888}{Michele Lombardi} \\
    Department of Computer Science and Engineering \\
    University of Bologna \\
    Viale del Risorgimento 2, Bologna (BO), Italy \\
    \texttt{michele.lombardi2@unibo.it} \\
}

\renewcommand{\headeright}{}
\date{}

\begin{document}

\maketitle

\begin{abstract}
The Hirschfeld–Gebelein–R{\'e}nyi (HGR) correlation coefficient is an extension of Pearson's correlation that is not limited to linear correlations, with potential applications in algorithmic fairness, scientific analysis, and causal discovery.
Recently, novel algorithms to estimate HGR in a differentiable manner have been proposed to facilitate its use as a loss regularizer in constrained machine learning applications.
However, the inherent uncomputability of HGR requires a bias-variance trade-off, which can possibly compromise the robustness of the proposed methods, hence raising technical concerns if applied in real-world scenarios.
We introduce a novel computational approach for HGR that relies on user-configurable polynomial kernels, offering greater robustness compared to previous methods and featuring a faster yet almost equally effective restriction. 
Our approach provides significant advantages in terms of robustness and determinism, making it a more reliable option for real-world applications. 
Moreover, we present a brief experimental analysis to validate the applicability of our approach within a constrained machine learning framework, showing that its computation yields an insightful subgradient that can serve as a loss regularizer.
\end{abstract}

\maketitle

\section{Introduction}

Detecting correlations is a recurring task in statistics and machine learning (ML), forming the foundation of numerous algorithms and applications.
Nevertheless, the most well-known indicators are restricted to specific types of correlation -- linear in the case of Pearson's coefficient, or monotonic with Spearman's rank.
A notable exception is the Hirschfeld–Gebelein–R{\'e}nyi (HGR) correlation coefficient~\citep{Rnyi1959OnMO}, which extends Pearson's coefficient to capture non-linear effects by means of two mapping functions on the input data, known as \textit{copula transformations}.

Several computational techniques have been suggested over the years to estimate the value of HGR, whose exact value is theoretically uncomputable, all requiring a certain balance between bias and variance in the estimation models to produce an accurate approximation.
Furthermore, most of these methods have relied on iterative processes and could not offer any gradient information about the computed outcome, thus hindering their use as loss regularizers for neural networks.
Inspired by the concept of using general correlation indicators as fairness metrics, recent studies from \citet{pmlr-v97-mary19a} and \citet{ijcai2020p313} have investigated differentiable algorithms for computing HGR with the goal of employing it as a loss regularizer.
However, the complexity of HGR raises concerns about the robustness of these methods, which may be particularly problematic in ethical or legal contexts.

In this work, we provide two main contributions.
First, we identify limitations in existing methods that significantly reduce their applicability due to their sensitivity to sampling noise and their non-deterministic behaviour.
Second, we propose a novel computational approach for HGR, which relies on user-configurable polynomial kernels and is more robust than the previous counterparts, along with a faster but almost equally performative restriction.
Our theoretical results are complemented with experimental analyses conducted on synthetic and real-world datasets, focusing on both detection and enforcement of correlations.

\section{Background and State of the Art}

There exist several indicators to compute the correlation between two variables.
Among the most straightforward and widely used are Pearson's correlation coefficient and Spearman's rank correlation.
Although very simple to interpret and compute, these two metrics are limited to measuring linear and monotonic correlations, respectively.
A natural but less renowned extension is the Hirschfeld–Gebelein–R{\'e}nyi (HGR) correlation coefficient, also known as maximal correlation coefficient.
It is defined as the maximal Pearson's correlation achievable by mapping the variables into non-linear spaces by means of copula transformations.
Formally, given two variables $A$ and $B$, we have:
\begin{equation}
    \label{eq:hgr}
    \mbox{HGR}(A, B) = \sup_{f, g \in \mathcal{F}} \rho(f(A), g(B))
\end{equation}
where $\rho(X, Y) = \frac{\operatorname{cov}(X, Y)}{\sigma(X) \sigma(Y)}$ is the Pearson's coefficient, and $f$ and $g$ are the two copula transformations belonging to the Hilbert space $\mathcal{F}$ of all the possible functions.
Notably, such definition allows to derive three important properties:
\begin{equation}
    \begin{split}
        \operatorname{HGR}(A, B) &\in \left[0, 1\right] \\
        \operatorname{HGR}(A, B) &= 1 \iff \exists f, g \mid P(f(A) = g(B)) = 1 \\
        \operatorname{HGR}(A, B) &= 0 \iff A \ci B
    \end{split}
\end{equation}
in other words, the domain of HGR is bounded between $0$ and $1$, reaching its peak when there exist two deterministic functions $f$ and $g$ such that the random variables become identical, and hitting its lowest point when $A$ and $B$ are independent. This last feature is particularly significant, as other correlation measures do not ensure it; for instance, two variables could be dependent without a linear relationship, resulting in a Pearson's correlation of zero.

\subsection{Algorithms for HGR Estimation}

Despite its benefits, HGR is hardly used in practice due to its need to optimize over an infinite set of infinite-dimensional elements -- i.e., all possible $f$ and $g$ functions.
Among the tractable approximations that have been proposed, the Alternating Conditional Expectations algorithm~\citep{doi:10.1080/01621459.1985.10478157} was the first to produce an estimate of HGR.
Similarly, other measures such as Distance and Brownian Correlation~\citep{Sz_kely_2009}, Kernel Independent Component Analysis~\citep{Bach_2003}, Kernel Canonical Correlation Analysis~\citep{Hardoon_2008}, and Hilbert-Schmidt Independence Criterion~\citep{10.5555/1046920.1194914,10.5555/3042573.3042782} have been developed to address comparable objectives.
Finally, the Randomized Dependence Coefficient~\citep{NIPS2013_aab32389} selects the highest correlated pair among randomly-calibrated sinusoidal projections of the input variables into a non-linear space.

The common backbone of this extensive literature lies in the idea of pairing the expressiveness guaranteed by non-linear kernel operations with the well-understood theoretical and practical advantages of linear algebra.
Nonetheless, the aforementioned approaches are uniquely designed to support correlation \textit{detection}, neglecting any possibility to \textit{enforce} a desired correlation value within a gradient-based learning environment.
As a solution, \citet{pmlr-v97-mary19a} develops a differentiable algorithm that estimates HGR according to a tractable upper bound known as Withsenhausen's characterization~\citep{Witsenhausen_1975}, while \citet{ijcai2020p313} further extends this work by proposing a novel method where the copula transformations are approximated by two adversarial neural networks.
Both these approaches provide meaningful gradients or sub-gradients, and can thus be effectively used as loss penalizers during the training procedure as proved by the reported experimental analysis involving fairness usecases.

\subsection{Limitations of Existing Approaches}

\begin{figure}
    \centering
    \includegraphics[width=0.7\textwidth]{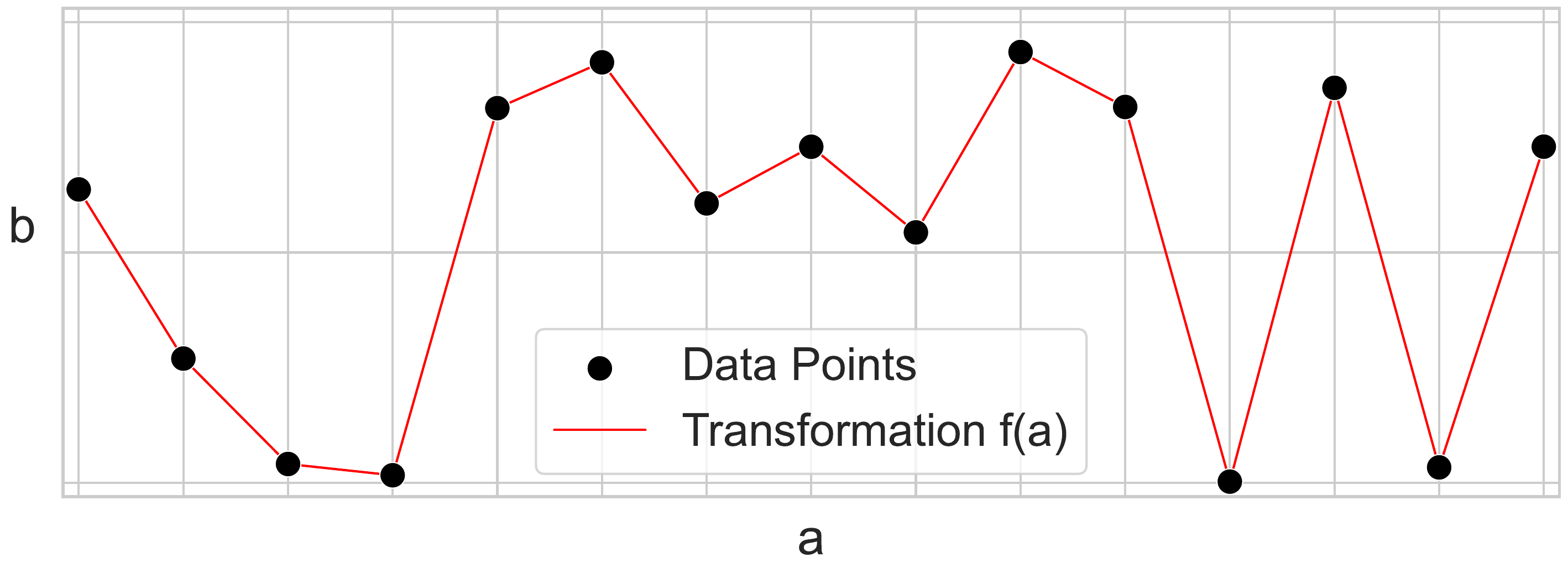}
    \caption{Example of overfitting in the computation of sample HGR}
    \label{fig:overfitting}
\end{figure}

Other than its uncomputability, another issue with HGR is that it is naturally defined for full distributions rather than finite samples.
In principle, a sample version of HGR can be easily obtained by swapping the sample Pearson's correlation with its theoretical value in \Cref{eq:hgr}.
However, as highlighted both by \citet{NIPS2013_aab32389} and \citet{pmlr-v202-giuliani23a}, this makes the indicator prone to overfitting on any sample $(x, y) \sim P(XY)$ whose pairs can be interpreted as the case-by-case specification of either a $x \mapsto y$ or $y \mapsto x$ function.

\Cref{fig:overfitting} shows how this task can be easily accomplished using a piecewise-linear function that precisely fits the data, but we underline that this concept applies to any model capable of exact interpolation on a dataset, ranging from piecewise-constant to polynomial or spline interpolations. In real-world scenarios, tackling these issues requires the use of sub-optimal computational methods that can balance the trade-off between bias and variance.
In practice, addressing this challenge involves using sub-optimal strategies able to balance their bias-variance trade-off.

For this purpose, the Randomized Dependence Coefficient (RDC) employs a fixed number of random sinusoidal copula transformations, later selecting the pair that maximizes co-linearity.
This approach retains good interpretability, since the chosen transformations can be easily plotted and analyzed.
However, randomization provides limited control of the bias-variance trade-off and makes the resulting approach non-deterministic, which is counterintuitive and might be detrimental in terms of user trust.
The method of \citeauthor{pmlr-v97-mary19a}, instead, relies on kernel-density techniques to estimate the probability distributions from which the vectors are sampled, ultimately computing the correlation according to a tractable HGR upper bound~\citep{Witsenhausen_1975}.
This approach is deterministic, but the use of a bound together with an approximation can lead to inaccurate results.
Moreover, the process relies on a discretization whose properties affect the final results in a complex way, and the $f$ and $g$ transformations are only implicitly defined, making them impossible to plot.
Such a combination makes this approach particularly opaque.
As regards the work by \citeauthor{ijcai2020p313}, it tackles the issue of uncomputability by relying on two Neural Networks (NNs), which are jointly trained in an adversarial manner to approximate the copula transformations.
This approach retains some interpretability, since the transformations can be plotted; yet, they cannot be easily analyzed due to the subsymbolic nature of the transformation.
Furthermore, the high expressivity of NNs proves to be a double-edge sword here.
Indeed, if on the one hand they avoid the need to commit to a class of functions, on the other hand they increase the risk of overfitting.
Finally, the computational process is based on Stochastic Gradient Descent, which tends to be significantly slower than the alternatives and leads to non-deterministic results.

\section{A Practical HGR Approximation}

The core idea of our approach is to represent the $f$ and $g$ functions by means of finite-degree polynomials.
Formally, let us consider two vectors $(a, b) \sim P(AB)$ sampled from the joint distribution of $A$ and $B$.
Then, our finite variance models take the form of weighted polynomial expansions $\mathbf{P}_x^d \cdot \omega$, where $x$ is the input vector, $d$ is the degree of the polynomial kernel, and $\omega$ is a $d$-dimensional vector of weights associated with each polynomial degree.
Specifically:
\begin{equation}
    \mathbf{P}_x^d \cdot \omega = \begin{pmatrix}
        x_1 & x_1^2 & \hdots & x_1^d \\
        x_2 & x_2^2 & \hdots & x_2^d \\
        \vdots & \vdots & \ddots & \vdots \\
        x_n & x_n^2 & \hdots & x_n^d \\
    \end{pmatrix} \cdot \begin{pmatrix}
        \omega_1 \\ \omega_2 \\ \vdots \\ \omega_d
    \end{pmatrix}
\end{equation}
Accordingly, we define our kernel-based HGR variant as:
\begin{equation}
    \label{eq:hgr_kb}
    \operatorname{HGR-KB}(a, b; h, k) = \max_{\alpha, \beta} \rho(\mathbf{P}_a^h \cdot \alpha, \mathbf{P}_b^k \cdot \beta)
\end{equation}
where the copula transformations $f(a)$ and $g(b)$ are substituted with $\mathbf{P}_a^h \cdot \alpha$ and $\mathbf{P}_b^k \cdot \beta$, respectively.
In this context, $h$ and $k$ are two positive integers that represent the order of polynomial expansions for both variables.
These hyperparameters, whose specification is designated to the user, offer a means to control the indicator's degree of freedom, both in terms of bias-variance trade-off and in terms of expressiveness versus higher computational demands.
The remainder of this section is focused on technical issues concerning the computation of the indicator, with the last part being instead dedicated to the discussion of the properties and applicability of the approach.

\subsection{Technical Analysis}

Addressing \Cref{eq:hgr_kb} is difficult, since the classical expression for Pearson's coefficient contains many non-linearities.
However, as shown in \Cref{app:pearson_lstsq}, Pearson's coefficient can be reformulated in terms of least-squares, leading to the following bi-level optimization definition for $\operatorname{HGR}(a, b)$:
\begin{equation}
    \label{eq:hgr_bilevel}
    \max_{f, g} \argmin_r \left\| \frac{f(a) - \mu(f(a))}{\sigma(f(a))} \cdot r\ - \frac{g(b) - \mu(g(b))}{\sigma(g(b))} \right\|_2^2
\end{equation}
This is an alternative yet equivalent formulation of the sample version of \Cref{eq:hgr}, from which the correlation can be retrieved as the optimal $r^*$ value provided that the copula transformations have finite and strictly positive variance.
Furthermore, in \Cref{app:hgr_single_level} we prove the alignment of the two objectives, allowing to cast the problem as a single-level optimization.
By plugging our polynomial models in place of the copula transformations, we obtain:
\begin{equation}
    \argmin_{\alpha, \beta, r} \left\| \frac{\mathbf{P}_a^h \cdot \alpha - \mu(\mathbf{P}_a^h \cdot \alpha)}{\sigma(\mathbf{P}_a^h \cdot \alpha)} \cdot r\ - \frac{\mathbf{P}_b^k \cdot \beta - \mu(\mathbf{P}_b^k \cdot \beta)}{\sigma(\mathbf{P}_b^k \cdot \beta)} \right\|_2^2
\end{equation}
Observing this equation reveals two main insights. First, the mean operator is translation-invariant, rendering the terms $\mu(\cdot)$ negligible since we can pre-compute the zero-centered polynomial kernels $\widetilde{\mathbf{P}}$. Second, the value $r$ is multiplied by a term that is scale invariant, due to the appearance of the standard deviation $\sigma(\mathbf{P}_a^h \cdot \alpha)$ in the denominator. As a consequence, both degrees of freedom can be merged by defining $\widetilde{\alpha} = \alpha r \cdot \sigma(\mathbf{P}_a^h \cdot \alpha)^{-1}$, resulting in:
\begin{gather}
    \label{eq:hgr_pearson}
    \operatorname{HGR-KB}(a, b; h, k) = \rho(\mathbf{P}_a^h \cdot \widetilde{\alpha}^*, \mathbf{P}_b^k \cdot \beta^*) \\
    \label{eq:hgr_optimization}
    \widetilde{\alpha}^*, \beta^* = \argmax_{\widetilde{\alpha}, \beta} \left\| \widetilde{\mathbf{P}}_a^h \cdot \widetilde{\alpha} - \frac{\widetilde{\mathbf{P}}_b^k \cdot \beta}{\sigma(\widetilde{\mathbf{P}}_b^k \cdot \beta)} \right\|_2^2
\end{gather}
\begin{figure}
    \centering
    \includegraphics[width=0.8\textwidth]{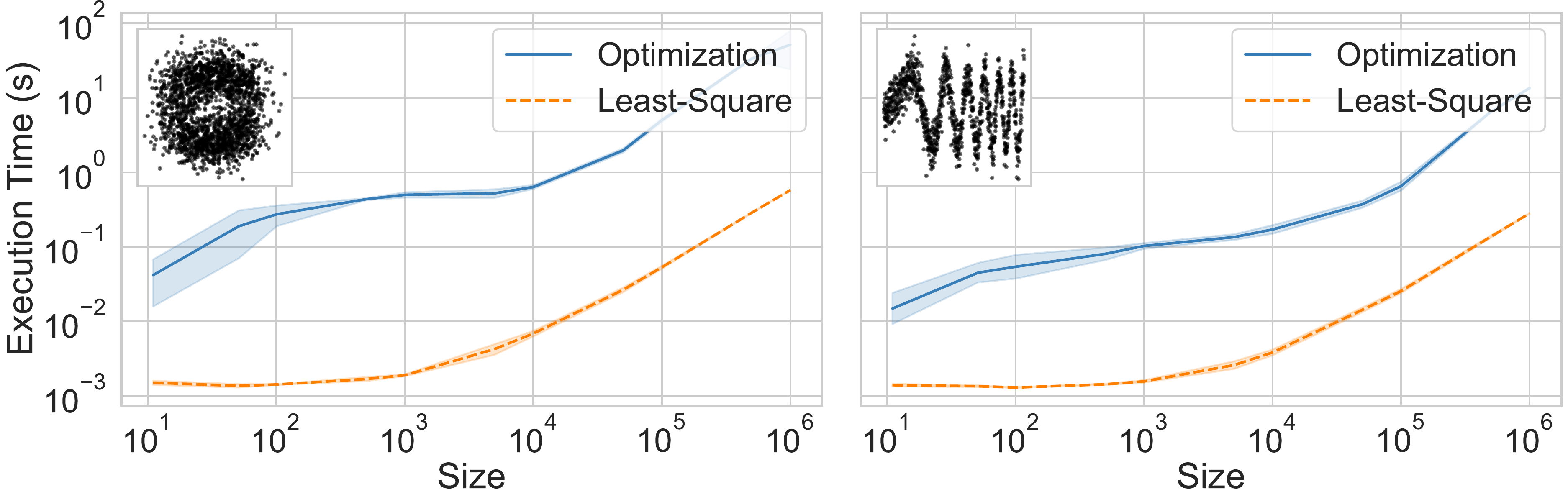}
    \caption{Time required to compute $\operatorname{HGR-SK}$ using Least-Square vs. Global Optimization algorithms.}
    \label{fig:lstsq}
\end{figure}
Unlike the original \Cref{eq:hgr_kb}, this formulation features a single non linearity -- the $\sigma(\widetilde{\mathbf{P}}_b^k \cdot \beta)$ denominator -- in addition to the least-square objective, making it much easier to address.
Moreover, rescaling the $\beta$ vector in \Cref{eq:hgr_optimization} does not change the value of the cost function, meaning that the problem admits infinitely many equivalent solutions.
Such symmetries can be removed by arbitrarily picking a value for the term $\sigma(\widetilde{\mathbf{P}}_b^k \cdot \beta)$.
For simplicity, we select $1$ and impose a constraint on the variance rather than the standard deviation, thus leading to:
\begin{align}
    \label{eq:hgr_constrained}
    \argmax_{\widetilde{\alpha}, \beta} \left\| \widetilde{\mathbf{P}}_a^h \cdot \widetilde{\alpha} - \widetilde{\mathbf{P}}_b^k \cdot \beta \right\|_2^2 \qquad \text{s.t. }  \sigma(\widetilde{\mathbf{P}}_b^k \cdot \beta)^2 = 1
\end{align}
Despite the simple quadratic objective, \Cref{eq:hgr_constrained} is not trivial to solve due to the presence of an equality constraint defined over the quadratic function $\sigma(\widetilde{\mathbf{P}}_b^k \cdot \beta)^2$.
However, a convex formulation can be obtained via careful application of Lagrangian methods, as proved in \Cref{app:lagrangian_soundness}, thus implying that a globally optimal solution exists.
In our implementation, we achieve it by addressing \Cref{eq:hgr_constrained} directly via one of the Trust Region Methods from \citet{Conn_2000}, specifically using the implementation provided in the \texttt{scipy.optimize} package.

\subsection{Single-Kernel Subcase}

When a polynomial of order 1 is used for one of the two kernels, \Cref{eq:hgr_constrained} reduces to:
\begin{align}
    \label{eq:hgr_leastsquares}
    \argmax_{\widetilde{\alpha}, \beta} \left\| \widetilde{\mathbf{P}}_a^h \cdot \widetilde{\alpha} - \beta (b - \mu(b)) \right\|_2^2 \qquad \text{s.t. } \sigma(\beta b) = 1
\end{align}
where the value of $\beta$ is completely determined by the constraint and equal to 1 / $\sigma(b)$.
As a consequence, \Cref{eq:hgr_leastsquares} is a classical least-square problem that can be solved very efficiently via any suitable method.
The main drawback of this setup is that it can only quantify correlations in functional form, e.g., $B \simeq f(A)$.
Still, the computational advantages are large enough that we chose to use it as the basis for a restricted version of our indicator, which we call Single-Kernel HGR.
$\operatorname{HGR-SK}$ is obtained by evaluating $\operatorname{HGR-KB}$ with orders $d, 1$ and $1, d$, then taking the largest result:
\begin{equation}
    \begin{split}
        \operatorname{HGR-SK}&(a, b; d) = \max \{ \rho(\widetilde{\mathbf{P}}_a^d \! \cdot \! \widetilde{\alpha}^*, b), \rho(a, \widetilde{\mathbf{P}}_b^d \! \cdot \! \widetilde{\beta}^*) \} \\
        \widetilde{\alpha}^* &= \argmax_{\widetilde{\alpha}} \left\| \widetilde{\mathbf{P}}_a^d \cdot \widetilde{\alpha} - \frac{b - \mu(b)}{\sigma(b)} \right\|_2^2 \\
        \widetilde{\beta}^* &= \argmax_{\widetilde{\beta}} \left\| \widetilde{\mathbf{P}}_b^d \cdot \widetilde{\beta} - \frac{a - \mu(a)}{\sigma(a)} \right\|_2^2
    \end{split}
\end{equation}
Here, $d$ controls the degree of both polynomial expansions, and the indicator can account for functional dependencies in both directions, i.e., $B \simeq f(A)$ and $A \simeq f(B)$, leaving out only cases of strong non-linear co-dependency.

The primary strength of the Single-Kernel formulation lies in its speed.
Solving an unconstrained least-square problem is a well-understood task in linear algebra, with highly-optimized computational routines available. 
\Cref{fig:lstsq} demonstrates that employing least-square solvers offers an improvement of nearly two orders of magnitude over global optimization using trust region methods.
Similar conclusions can be drawn from the experiments shown in \Cref{sub:correlations}. 

\subsection{Properties and Applicability}

\begin{table*}
    \centering
    \begin{tabular}{c|ccccc}
        \toprule
        Method & $\operatorname{HGR-KB}$ & $\operatorname{HGR-SK}$ & $\operatorname{HGR-NN}$ & $\operatorname{HGR-KDE}$ & $\operatorname{RDC}$ \\
        \midrule
        Expressivity & $f$, $g$ & $f$ \textbf{or} $g$ & $f$, $g$ & $f$, $g$ (distributions) & $f$, $g$ \\
        Interpretability & $\checkmark$ & $\checkmark$ & visualization only & $\times$ & $\checkmark$ \\
        Configurability & $\checkmark$ & $\checkmark$ & architecture only & $\times$ & $\times$ \\
        Differentiability & $\checkmark$ & $\checkmark$ & $\checkmark$ & $\checkmark$ & $\times$ \\
        Determinism & $\checkmark$ & $\checkmark$ & $\times$ & $\checkmark$ & $\times$ \\
        \bottomrule
    \end{tabular}
    \caption{Properties of our methods ($\operatorname{HGR-KB}$ and $\operatorname{HGR-SK}$) compared to three alternative techniques for computing HGR.}
    \label{tab:properties}
\end{table*}

We argue that our indicators enjoy a number of properties that make them considerably better suited for real-world applications compared to alternatives.
\Cref{tab:properties} reports a summary of these properties.

\paragraph{Expressivity}
In terms of expressivity, both our approach and the adversarial one use universal approximators.
In principle, polynomials run into numerical issue much earlier than neural networks, but in practice very high expressivity is not necessarily desirable due to the risk overfitting, as our experiments will show.
Conversely, the KDE method relies on a bound, thus making an expressivity analysis complex to perform, while the RDC method makes use of a single sinusoidal function and is therefore strictly less expressive.

\begin{figure}
    \centering
    \includegraphics[width=0.72\textwidth]{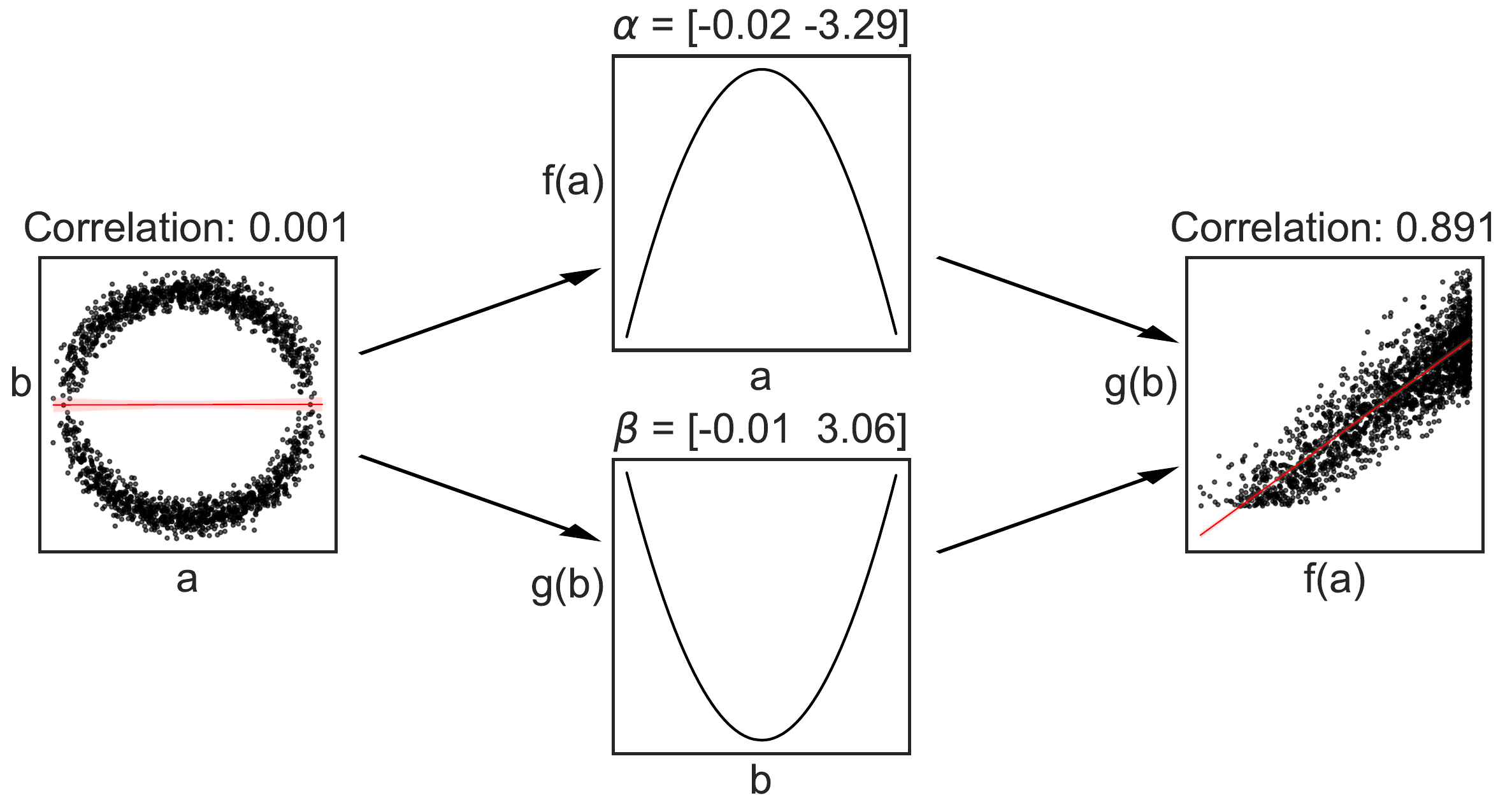}
    \caption{Example of kernel computation.}
    \label{fig:example}
\end{figure}

\paragraph{Interpretability}
The use of polynomial kernels makes our approach particularly easy to examine, on par with the Randomized Dependency Coefficient.
\Cref{fig:example} shows an example of how the optimized kernels can be plotted and analytically inspected in terms of their interpretable coefficients.
Original data is depicted on the left, showing almost no (linear) correlation.
The central figures showcase the learned copula transformations, along with the respective coefficients for the polynomial terms.
Finally, the projected data is plotted on the right, revealing a significantly stronger correlation.
Our method directly links the magnitude of each component to the degree it represents; for example, in the depicted case we can clearly discern the quadratic relationship between the variables by looking at the magnitude of the second and the first terms in the coefficient vectors.
This feature is absent in the kernel-density estimation method, while the adversarial method can only provide visualization due to the inherent sub-symbolic nature of neural networks.

\begin{figure*}[!b]
    \centering
    \subfigure[2015 US Census]{%
        \includegraphics[width=0.3\textwidth]{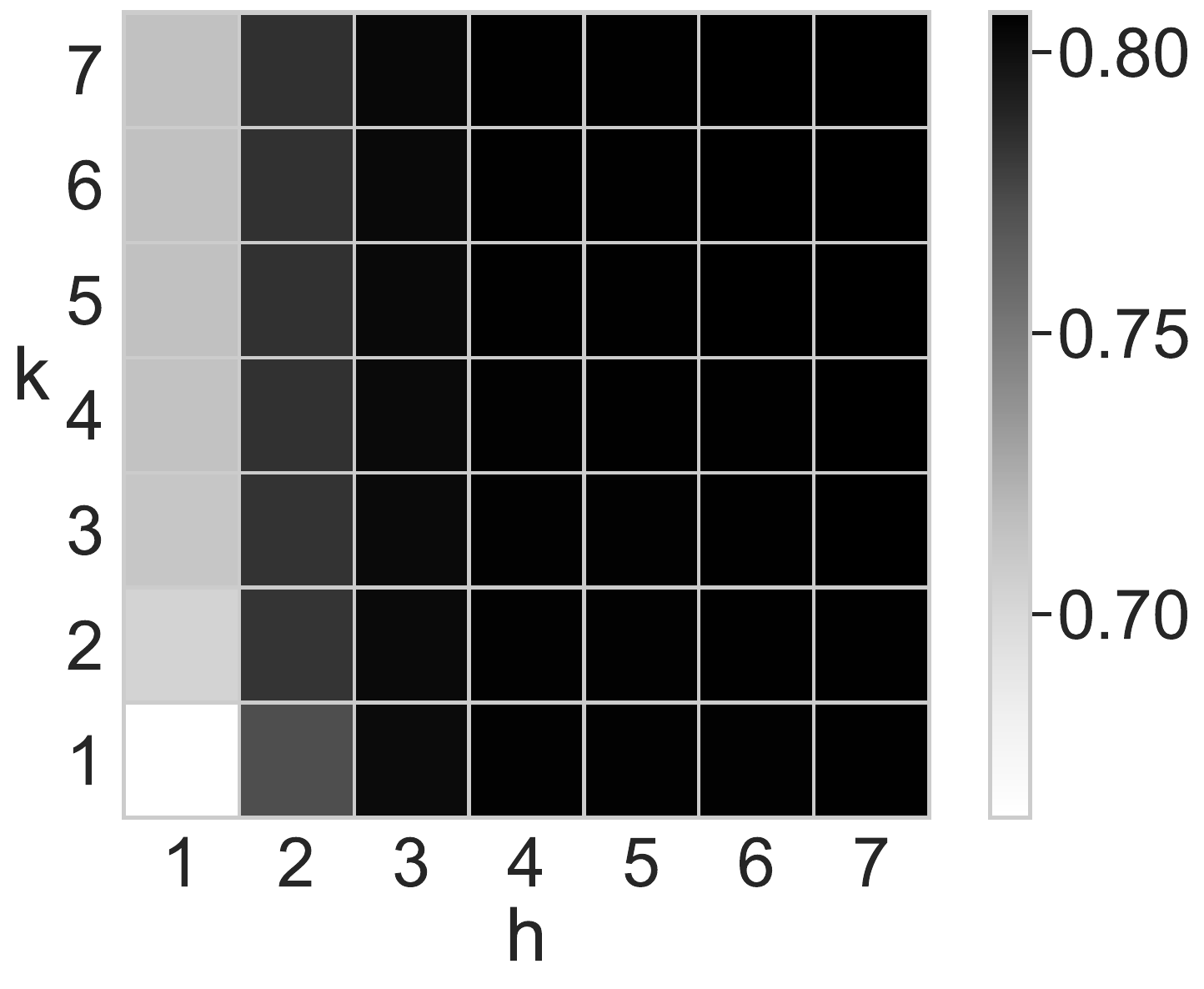}
    }\quad
    \subfigure[Communities \& Crimes]{%
        \includegraphics[width=0.3\textwidth]{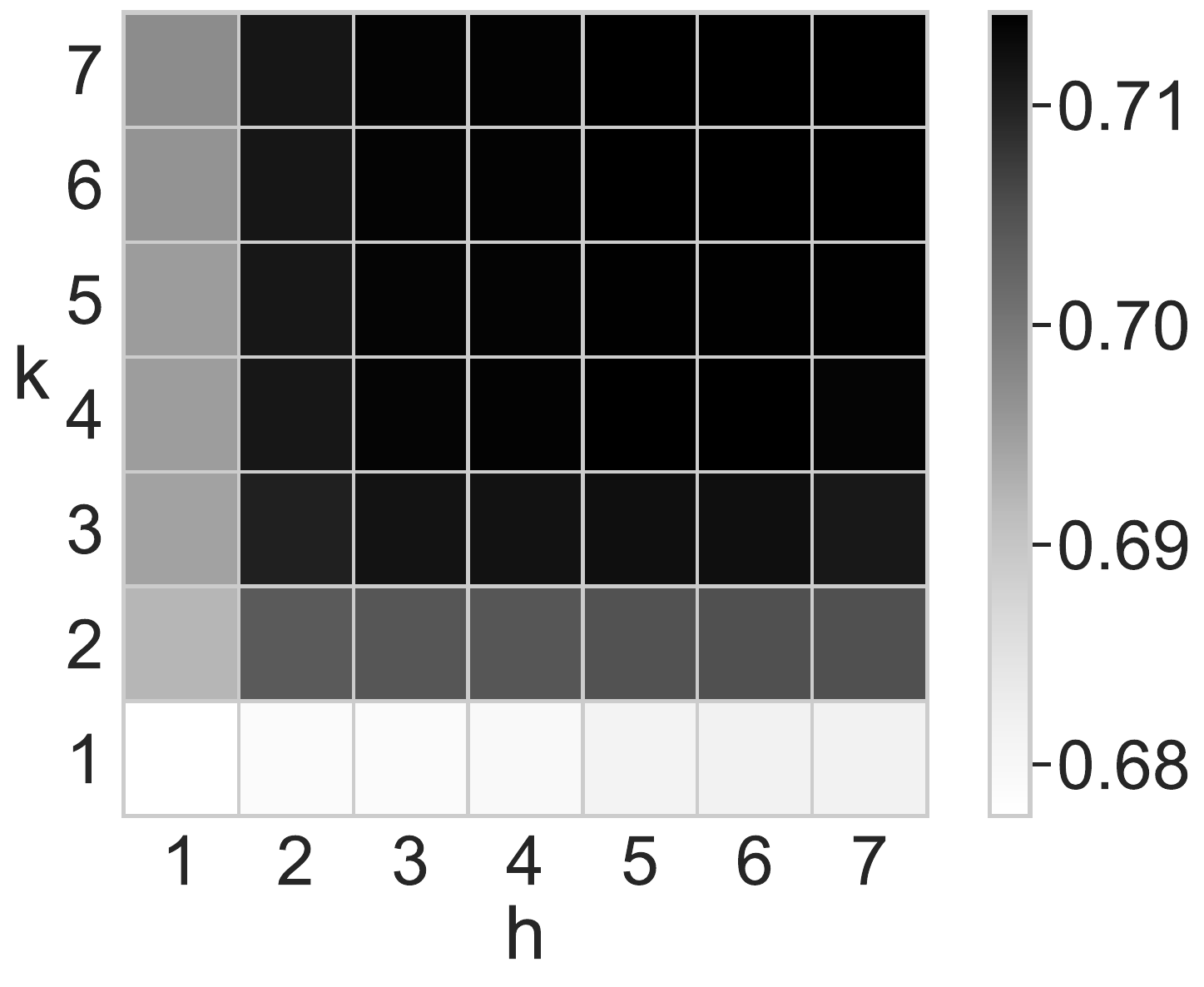}
    }\quad
    \subfigure[Adult Census Income]{%
        \includegraphics[width=0.3\textwidth]{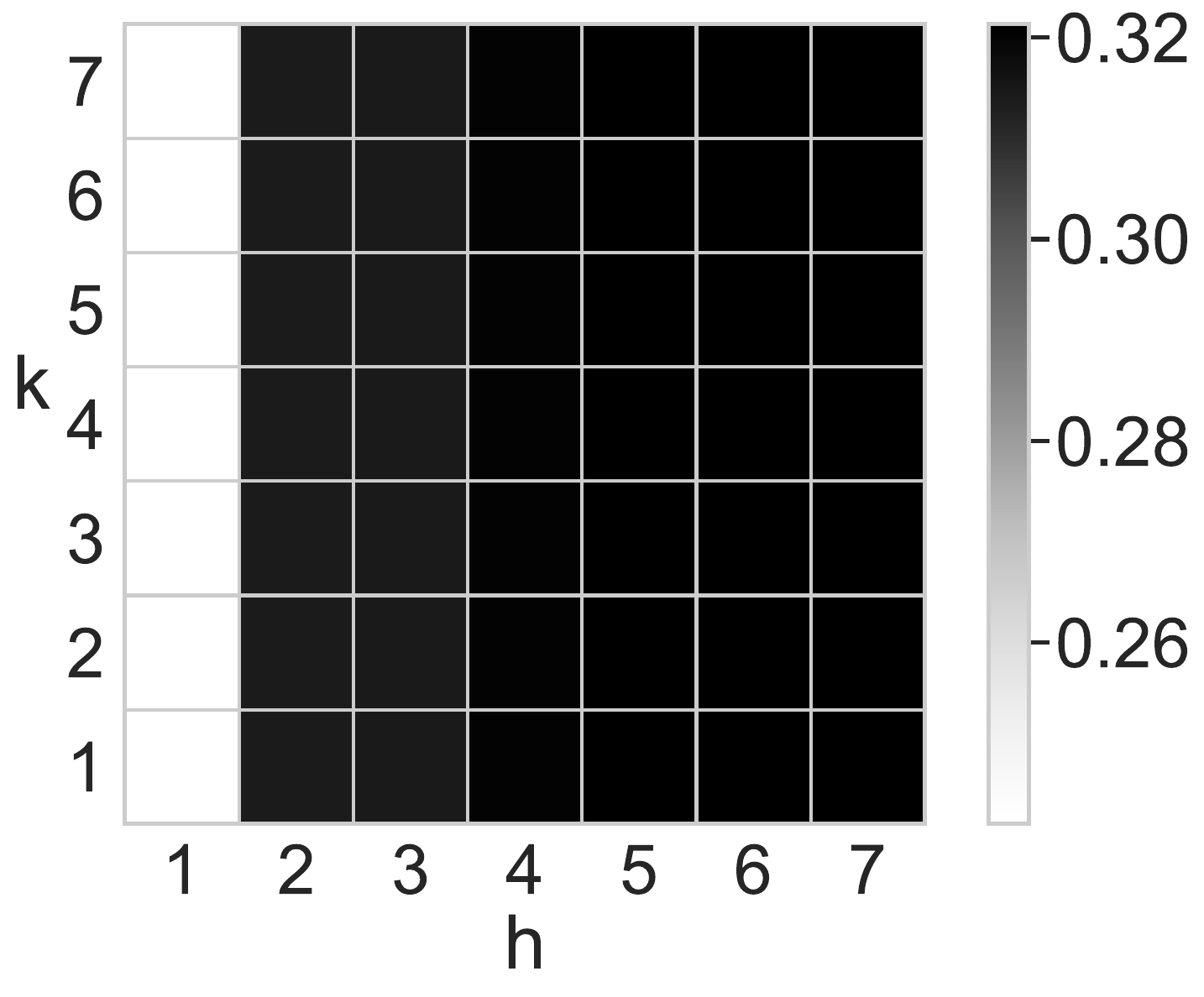}
    }
    \caption{Computation of $\operatorname{HGR-KB}(a, b; h, k)$  with varying $h$ and $k$ on three benchmark datasets.}
    \label{fig:monotonicity}
\end{figure*}

\paragraph{Configurability}
The ability to choose the kernel degrees provides a transparent and mathematically well-understood mechanism to control the bias-variance trade-off, which is critical for the method's robustness.
In \Cref{app:proof_monotonicity}, we prove that $\operatorname{HGR-KB}$ increases monotonically with respect to the degrees of the kernel, meaning that $\operatorname{HGR-KB}(a, b; h^\prime, k^\prime) \geq \operatorname{HGR-KB}(a, b; h, k)$ for any integer values $h^\prime \geq h, k^\prime \geq k$.
This property allows for a better understanding of the trade-off between bias and variance, as well as between expressiveness and computational requirements, by examining the improvement brought by higher degrees.
\Cref{fig:monotonicity} empirically validates this result on three benchmark datasets -- \textit{2015 US Census} ($a$~=~\texttt{Income}, $b$~=~\texttt{ChildPoverty}), \textit{Communities \& Crimes} ($a$~=~\texttt{pctWhite}, $b$~=~\texttt{violentPerPop}), and \textit{Adult} ($a$~=~\texttt{age}, $b$~=~\texttt{income}).
Darker colors indicate a higher correlation, thus proving the monotonically increasing relationship of the yielded correlation in both directions.
Moreover, it shows that the optimal trade-off for kernel degrees is likely to be found in $h, k \approx 3$, since lower values would result in an underestimation of the correlation, while higher values would bring little to no improvement.
It is worth mentioning that, although these parameters are data-specific, their calibration is much more intuitive and less demanding than the other methods.
In comparison, the parameters of the KDE approach have much less predictable effects, and only loose guidance can be provided to the RDC coefficient due to its randomness; as for the adversarial approach, it allows for a good degree of control, but through a less transparent mechanism due to the opaqueness of NNs.

\paragraph{Differentiability}
Many approaches for constrained machine learning are based on the use of regularizers to encourage the satisfaction of constraints at training time, which require differentiability of the chosen indicator.
The process of computing $\widetilde{\alpha}^*$ and $\beta^*$ from \Cref{eq:hgr_constrained} cannot be easily differentiated, since it relies on a numerical optimization procedure; however, \Cref{eq:hgr_pearson} is differentiable, meaning that $\operatorname{HGR-KB}(a, b; h, k)$ can yield a valid subgradient.
Moreover, the computation procedure for $\operatorname{HGR-SK}(a, b; d)$ is based on an unconstrained least-square problem, therefore it has a well-defined gradient since automatic differentiation frameworks support a differentiable least-squares operator, such as \texttt{tf.linalg.lstsq} in Tensorflow and \texttt{torch.linalg.lstsq} in PyTorch.

\begin{figure}
    \centering
    \subfigure[$\operatorname{HGR-KB}$]{%
        \includegraphics[width=0.27\textwidth]{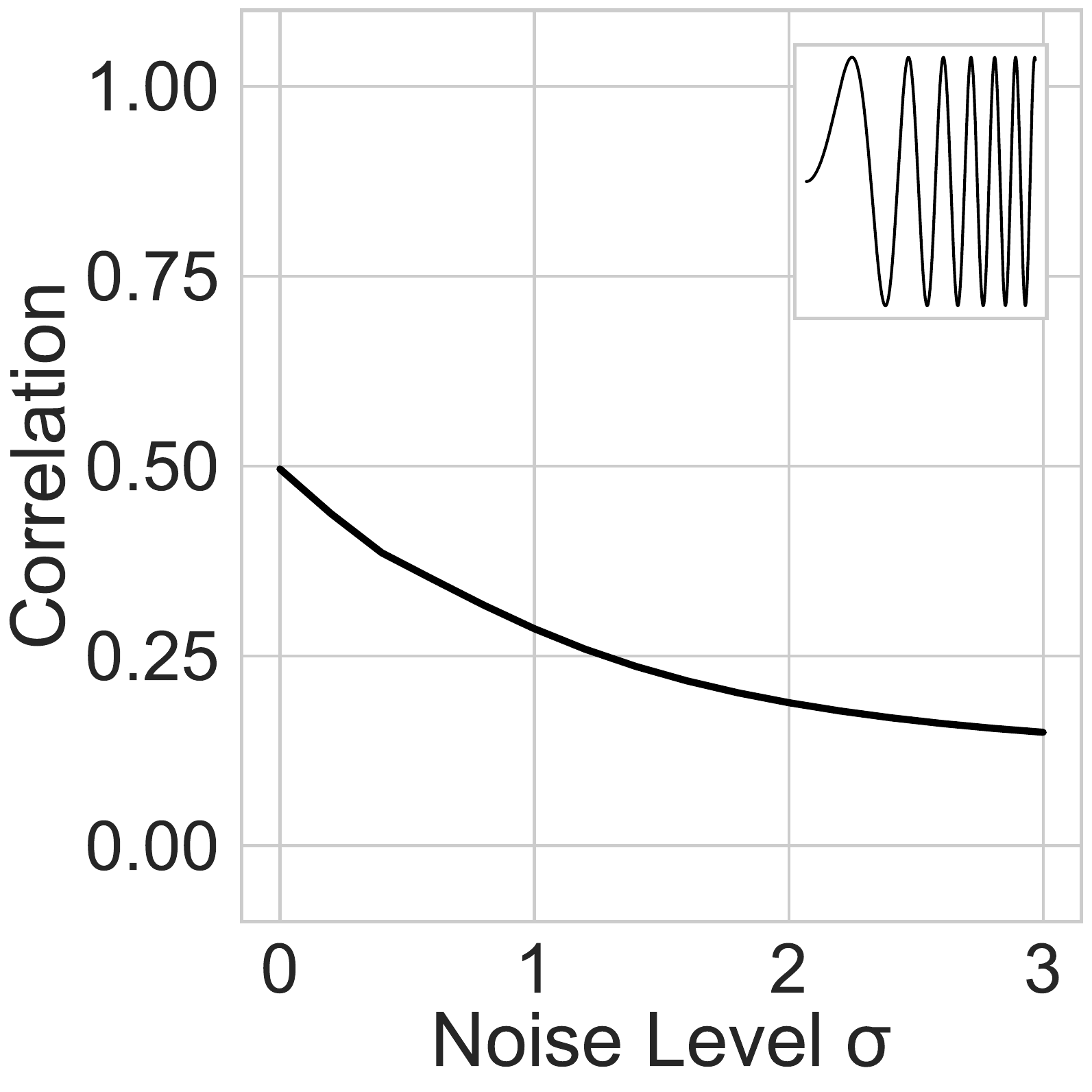}%
    }\qquad
    \subfigure[$\operatorname{HGR-NN}$]{%
        \includegraphics[width=0.27\textwidth]{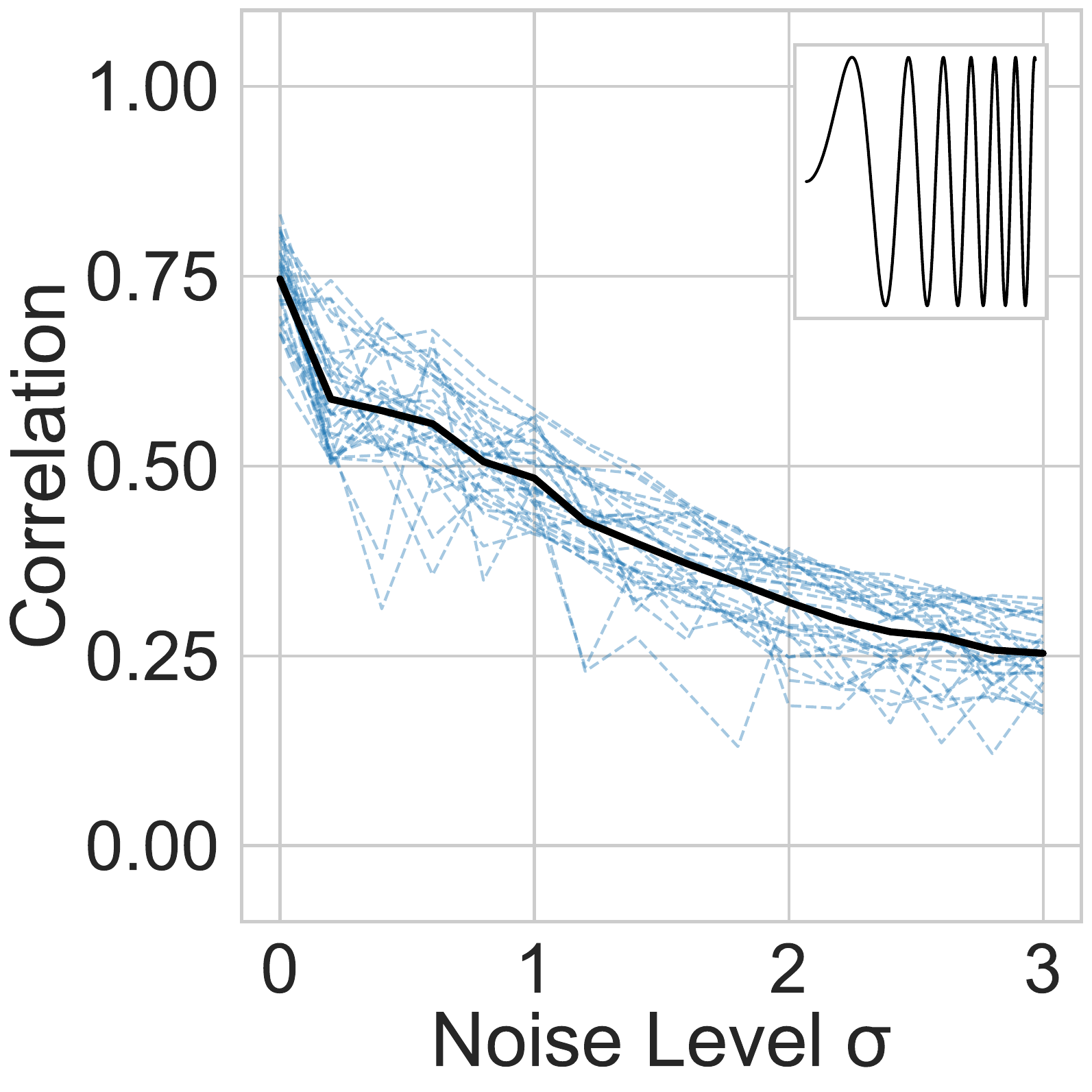}%
    }\qquad
    \subfigure[$\operatorname{RDC}$]{%
        \includegraphics[width=0.27\textwidth]{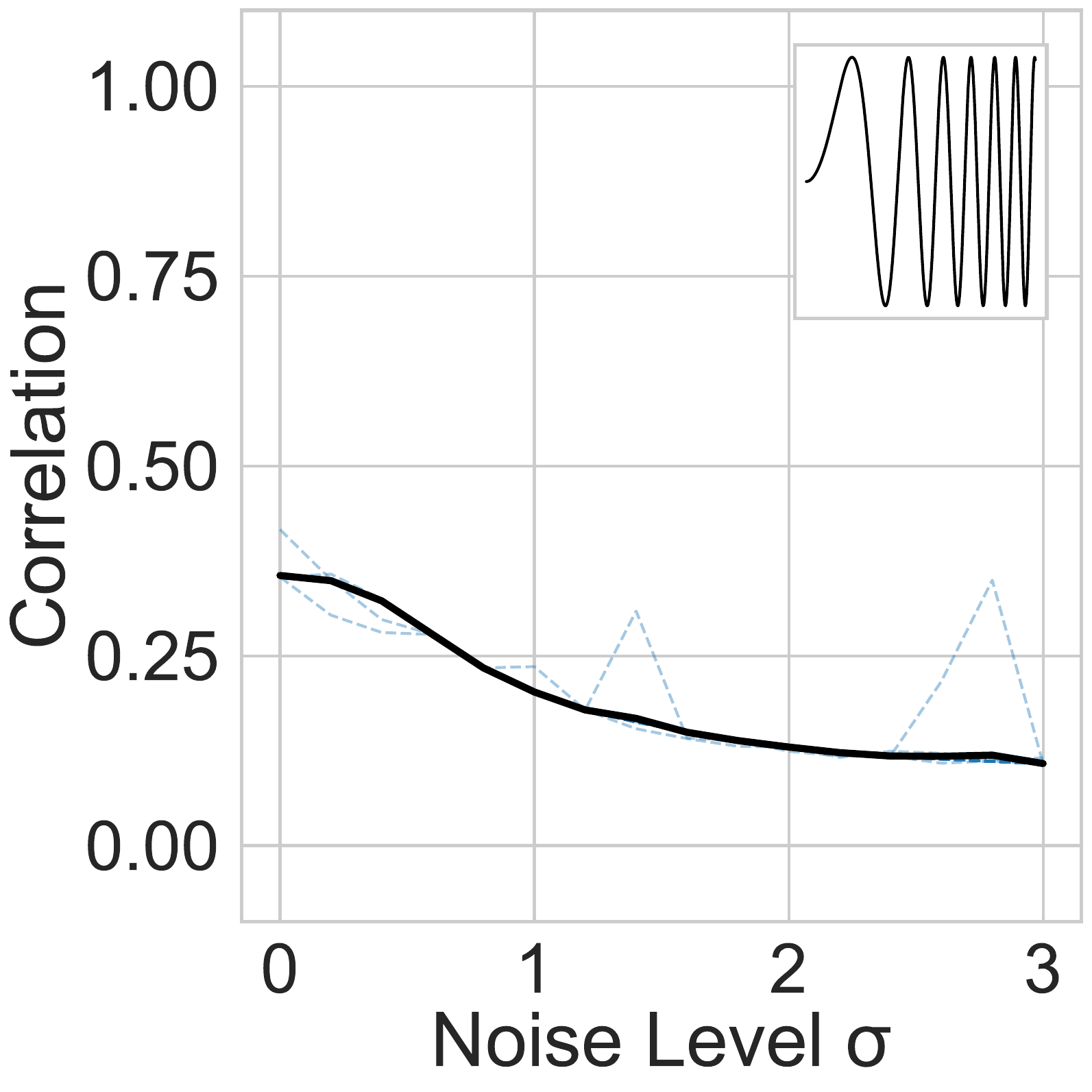}%
    }
    \caption{Effects of algorithm stochasticity in three different techniques to estimate HGR.}
    \label{fig:determinism}
\end{figure}

\paragraph{Determinism}
Using exact optimization methods makes our indicator fully deterministic, once the kernel degrees are specified.
This property is shared with the KDE approach, but not with RDC and the NN methods, which leverage intrinsically random operations -- random sinusoidal functions and Stochastic Gradient Descent.
Non-determinism can be a serious drawback in practical applications, as it may either cause confusion in decision-makers, or require multiple evaluations to get a robust measurement.
\Cref{fig:determinism} reports the impact of algorithmic stochasticity across three distinct HGR computation techniques, where the blue dashed lines represent single runs while the black solid line is the average.
As we can observe, our method produces consistent results for each of the $30$ seeds tested, whereas the outcomes of $\operatorname{HGR-NN}$ exhibit strong fluctuations while those of $\operatorname{RDC}$ exhibit few irregular peaks along with constant minor fluctuations.
We reinforce that non-determinism can be a significant disadvantage in real-world applications, as it may either lead to confusion among decision-makers or require multiple evaluations to achieve a reliable measurement, thus impacting the computational complexity.

\section{Experimental Analysis}

We will now discuss the specifics of our approach to demonstrate its strengths and limitations when used both as a correlation measure and as a loss penalizer.
We denote our two approaches as $\operatorname{HGR-KB}$ and $\operatorname{HGR-SK}$, with respective hyperparameters set to $h = k = 5$ and $d = 5$ unless otherwise stated.
The comparative baselines include the adversarial method from \citeauthor{ijcai2020p313} ($\operatorname{HGR-NN}$), the kernel-density one from \citeauthor{pmlr-v97-mary19a} ($\operatorname{HGR-KDE}$), and the Randomized Dependence Coefficient from \citeauthor{NIPS2013_aab32389} ($\operatorname{RDC}$).
We implemented our experiments in Python 3.10 and executed them on MacBook Pro with a 2.7GHz Intel Core I5 Dual-Core Processor and 8GB 1867 MHz DDR3 RAM -- no Graphics Processing Unit (GPU) is used and, before each run, we set a specific seed using the \texttt{seed\_everything} function of Pytorch Lightning, and use the \texttt{deterministic=True} training option to ensure reproducibility.
The code is publicly available at: \url{https://github.com/giuluck/kernel-based-hgr}.

\subsection{Correlation Detection on Synthetic Data}
\label{sub:correlations}

\begin{figure}
    \centering
    \includegraphics[width=0.85\textwidth]{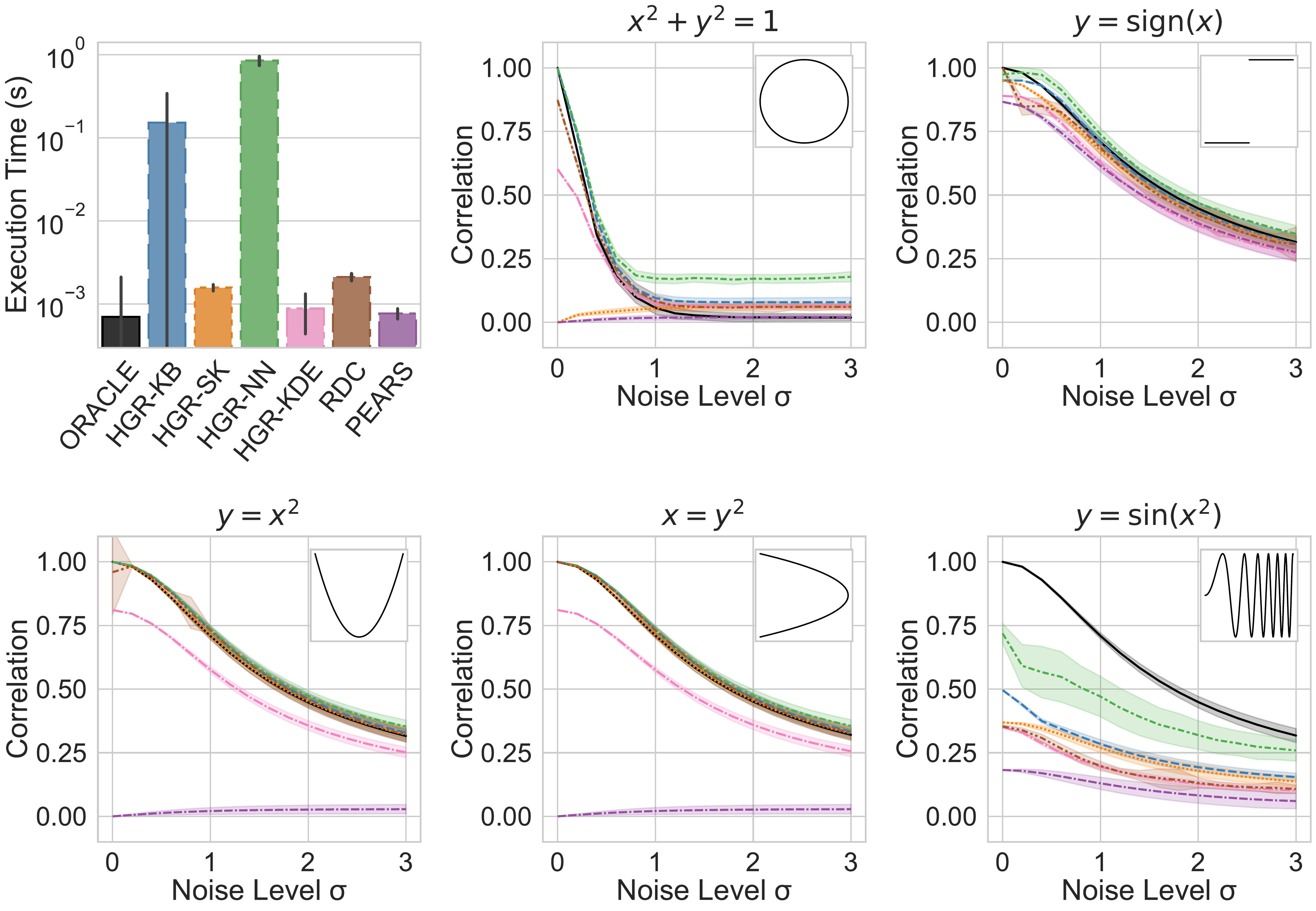}
    \caption{Execution times and correlations computed using several indicators across distinct deterministic relationships with varying degrees of noise.}
    \label{fig:correlations}
\end{figure}

\Cref{fig:correlations} illustrates the correlation computed by various indicators in a controlled setting where data were generated by adding Gaussian noise over pre-decided deterministic relationships.
The $\operatorname{ORACLE}$ method denotes the Pearson's correlation computed using optimal copula transformations -- e.g., $f(a) = a^2$ and $g(b) = -b^2$ for circular data.
For each function, we sample the noise vector using $10$ different seeds and run each method for $10$ different iterations.
The error bars represent the standard deviation of the obtained results, which originates from both stochasticity in the noise sampling and -- for non-deterministic approaches like $\operatorname{HGR-NN}$ and $\operatorname{RDC}$ -- the stochasticity in the solving process.

Most methods show similar performance, except for $\operatorname{HGR-KDE}$ which often underestimates the oracle value.
Regarding computational load, $\operatorname{HGR-SK}$ proves to be the fastest approach, although it fails to provide a good estimate in the circular dataset due to its inherent limitation to functional dependencies.
As expected, non-deterministic indicators -- $\operatorname{HGR-NN}$ and $\operatorname{RDC}$ -- exhibit higher variability while, at the expense of a higher cost, our method $\operatorname{HGR-KB}$ demonstrates better stability and produces reliable results, as evidenced by its proximity to the oracle in all cases except for the $y = \sin(x^2)$ example, where our chosen kernel orders were likely insufficient.

\begin{figure}
    \centering
    \includegraphics[width=0.85\textwidth]{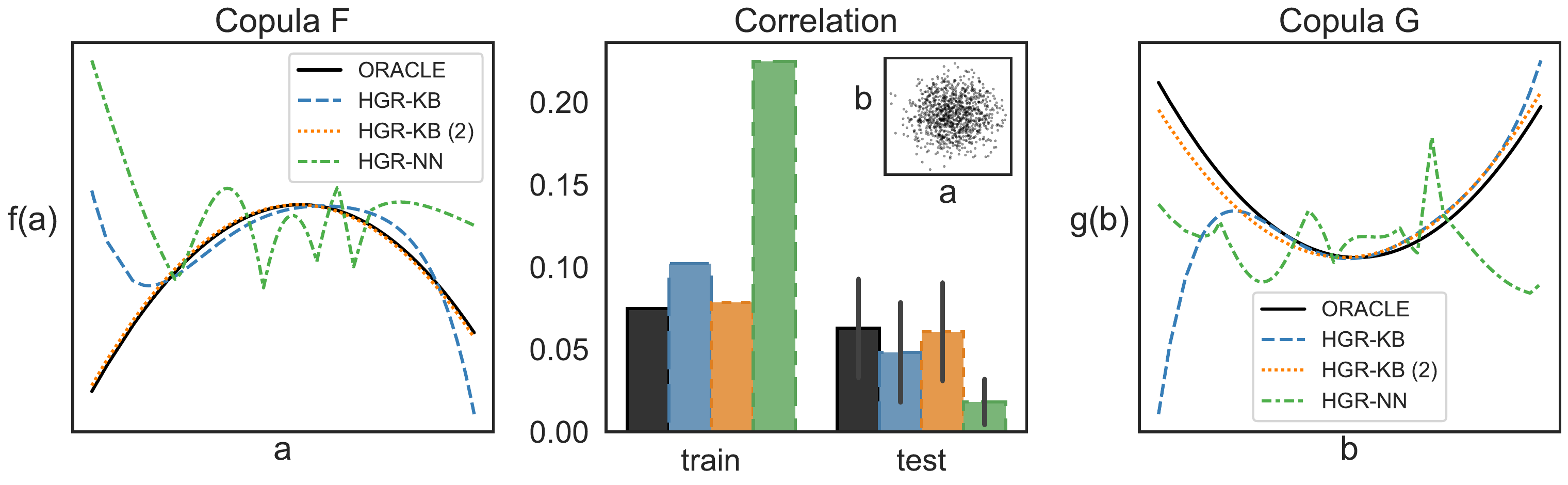}
    \caption{Kernel inspection of three HGR indicators on circular dataset.}
    \label{fig:copulas}
\end{figure}

Interestingly, the tendency of $\operatorname{HGR-NN}$ to overestimate the true correlation in the circular dataset can be most likely attributed to the remarkable expressivity of neural networks, which in this case leads to overfitting.
As a support for this claim, \Cref{fig:copulas} shows how the mappings generated by $\operatorname{HGR-NN}$ are considerably more unstable compared to our kernel-based counterparts.
More specifically, we consider the circular dependency with noise $\sigma = 1.0$ and examine the copula transformations generated by: (\textit{i}) our method $\operatorname{HGR-KB}$ with default hyperparameters $h = k = 5$, (\textit{ii}) a variant of our method $\operatorname{HGR-KB}$ ($2$) with hyperparameters $h = k = 2$, and (\textit{iii}) the adversarial method $\operatorname{HGR-NN}$.
Looking at the left ($f$) and right ($g$) plots, we observe how the neural transformations significantly overfit in certain regions, whereas our method tends to produce instability almost only at the borders.
Additionally, since our method allows to reduce the complexity of the transformations based on domain knowledge or experimental analysis, we could find that polynomial kernels of order 2 yield an even more stable result for this particular dataset, something that aligns in fact with the underlying deterministic relationship.
Consequently, the calculated correlations between the transformed variables do not hold when the learned kernels are used to compute correlations on different data points sampled from the same distribution -- ``test'' data in the figure --, as show by the poor results obtained on this test split, which do not reflect the training capabilities of the method.

\subsection{Correlation Enforcement in Fairness Scenarios}
\label{sub:benchmarks}

Our next question concerns the applicability of $\operatorname{HGR-KB}$ on constraint enforcement during the training of a Neural Network.
Specifically, we will tackle the case of unfairness penalization with respect to continuous protected attributes considering the $\operatorname{HGR-KB}$, $\operatorname{HGR-SK}$, and $\operatorname{HGR-NN}$ approaches only for comparison.
In the experiments, we constrain the correlation indicators to be lower than a certain threshold $\tau$ using the Lagrangian dual framework \citep{Fioretto2021}.
The approach requires a differentiable penalizer that, as previously discussed, yields a subgradient in the case of the $\operatorname{HGR-KB}$ indicator and an actual gradient for $\operatorname{HGR-SK}$.
A subgradient is also employed in $\operatorname{HGR-NN}$, as the training procedure of the adversarial networks must be detached from that of the predictive network, resulting in no information about its ``training'' gradients.

We evaluate our approach using three common benchmark datasets for fairness: US 2015 Census (\textit{Census}), Communities \& Crime (\textit{Communities}), and Adult Census Income (\textit{Adult}).
For each datasets, we select an output target ($y$) based on the task, and a protected attribute ($z$), namely a continuous sensitive feature correlated with the target.
Subsequently, we normalize the target variable, standardize the continuous inputs, one-hot encode multi-class inputs, and eventually train a fully-connected neural network while enforcing the fairness constraint up to a certain threshold $\tau$.
To do that, we employ the following custom loss:
\begin{equation}
    \mathcal{L}(\hat{y}(\theta), y) + \lambda \cdot \max \{0, \operatorname{HGR}(z, \hat{y}(\theta)) - \tau\}
\end{equation}
where $\mathcal{L}$ is the task loss, $\hat{y}(\theta)$ the model predictions, $y$ the ground truths, and $z$ the continuous protected attribute.
This formulation enables us to penalize any correlation exceeding a specific threshold $\tau$, which we define for each dataset as indicated in \Cref{tab:results}.
In order to guarantee constraint enforcement, we rely on the Lagrangian dual framework outlined in \citet{Fioretto2021}, as it automatically adjusts the weight $\lambda$ throughout the learning process via a gradient ascent step.
This requires setting an additional optimizer, which we define as Adam with learning rate $\operatorname{lr} = 10^{-3}$.
We opted for this method rather than leveraging a fixed weight $\lambda$ for our penalty as it eliminated the need for an additional tuning phase, enhancing efficiency and providing a better way to compare results.
For further details about this approach, we refer the reader to the original paper.

For each dataset, we run a $5$-fold cross-validation procedure.
As regards the methods, we amortize the run-time of both $\operatorname{HGR-KB}$ and $\operatorname{HGR-NN}$ by relying on warm starting.
In fact, when performing fairness enforcement in differentiable ML, we can use the information of the previous learning step to accelerate convergence in the next one.
In case of $\operatorname{HGR-KB}$, this is done by using the coefficients at gradient iteration $i$ as initial guesses for iteration $i + 1$, while for $\operatorname{HGR-NN}$ we adopt the original approach by \citeauthor{ijcai2020p313}, which fine-tunes the adversarial networks from the previous step for $50$ rather than $1000$ epochs.

\begin{table}
    \centering
    \adjustbox{width=\textwidth}{\begin{tabular}{l|c|cc|cc|c}
\toprule
\multicolumn{1}{c|}{Dataset} & Regularizer & \multicolumn{2}{c|}{Score} & \multicolumn{2}{c|}{Constraint} & Time (s) \\
& & train & val & train & val & \\
\midrule
\textsc{Census} & \textbf{//} & \textbf{0.70} ± \textbf{0.00} & \textbf{0.69} ± \textbf{0.00} & \textbf{//} & \textbf{//} & \textbf{32} ± \textbf{00} \\
\multirow{3}{*}{
    \footnotesize
    $\Bigg\{ \begin{array}{l}
        \tau = 0.4\\
        z = \texttt{Income}\\
        y = \texttt{ChildPoverty}
    \end{array}$
} & HGR-KB & 0.21 ± 0.03 & 0.20 ± 0.02 & 0.36 ± 0.03 & 0.36 ± 0.03 & 70 ± 03 \\
& HGR-SK & 0.23 ± 0.02 & 0.22 ± 0.01 & 0.36 ± 0.02 & 0.36 ± 0.02 & 38 ± 00 \\
& HGR-NN & 0.19 ± 0.04 & 0.19 ± 0.04 & 0.35 ± 0.05 & 0.35 ± 0.04 & 88 ± 00 \\
\midrule
\textsc{Communities} & \textbf{//} & \textbf{1.00} ± \textbf{0.00} & \textbf{0.52} ± \textbf{0.02} & \textbf{//} & \textbf{//} & \textbf{06} ± \textbf{00} \\
\multirow{3}{*}{
    \footnotesize
    $\Bigg\{ \begin{array}{l}
        \tau = 0.3\\
        z = \texttt{pctWhite}\\
        y = \texttt{violentPerPop}
    \end{array}$
} & HGR-KB & 0.74 ± 0.04 & 0.27 ± 0.05 & 0.28 ± 0.03 & 0.37 ± 0.07 & 39 ± 04 \\
& HGR-SK & 0.74 ± 0.05 & 0.27 ± 0.06 & 0.29 ± 0.02 & 0.36 ± 0.10 & 12 ± 00 \\
& HGR-NN & 0.72 ± 0.05 & 0.28 ± 0.07 & 0.30 ± 0.03 & 0.46 ± 0.07 & 60 ± 01 \\
\midrule
\textsc{Adult} & \textbf{//} & \textbf{0.92} ± \textbf{0.00} & \textbf{0.91} ± \textbf{0.00} & \textbf{//} & \textbf{//} & \textbf{15} ± \textbf{00} \\
\multirow{3}{*}{
    \footnotesize
    $\Bigg\{ \begin{array}{l}
        \tau = 0.2\\
        z = \texttt{age}\\
        y = \texttt{income}
    \end{array}$
} & HGR-KB & 0.88 ± 0.00 & 0.88 ± 0.01 & 0.19 ± 0.01 & 0.19 ± 0.01 & 57 ± 01 \\
& HGR-SK & 0.88 ± 0.00 & 0.87 ± 0.01 & 0.20 ± 0.01 & 0.20 ± 0.01 & 22 ± 00 \\
& HGR-NN & 0.88 ± 0.00 & 0.88 ± 0.00 & 0.19 ± 0.00 & 0.20 ± 0.00 & 73 ± 01 \\
\bottomrule
\end{tabular}}
    \caption{Results of experiments conducted on the benchmark datasets.}
    \label{tab:results}
\end{table}

\Cref{tab:results} presents the results of our investigation, where // indicates the unconstrained NN, while $\operatorname{HGR-*}$ denotes the loss penalizer.
For each run, we measure an accuracy score -- $R^2$ for \textit{Communities} and \textit{Census}, $\operatorname{AUC}$ for \textit{Adult} --, and the level of constraint satisfaction using the adopted penalizer between the continuous protected attribute and the predicted target.
In particular, the \textit{Constraint} column denotes the value of the enforced $\operatorname{HGR-*}$ penalty with respect to the continuous protected attribute.
Since different penalizers are employed, the enforced constraints are based on slightly different semantics; for this reason, we report $\operatorname{HGR}$ values for each of the penalizers being compared: constraint satisfaction should be checked for the matching $\operatorname{HGR}$ type, which is highlighted in bold font.

The results demonstrate our ability to effectively enforce constraints at -- or close to -- the desired level in each scenario.
Accuracy scores are comparable across all the tested penalizers, with $\operatorname{HGR-NN}$ often performing a bit worse, most likely due its tendency to overestimate the true correlation.
We remark that the reported constraint satisfaction is measured using the adopted penalizer, hence no direct accuracy comparison among the methods can be done as we have no access to an oracle able to yield the actual correlation between the continuous protected input and the output target.
In terms of training time, $\operatorname{HGR-SK}$ is significantly faster than the others at the expense of a slight increase in unfairness, as measured by the more reliable $\operatorname{HGR-KB}$ indicator.

\section{Conclusions}

We presented a novel methodology for computing the Hirschfeld-Gebelein-R{\'e}nyi ($\operatorname{HGR}$) correlation coefficient employing two polynomial kernels as copula transformations.
Our approach offers distinct advantages in terms of robustness, interpretability, and determinism, making it a more suitable option for real-world fairness scenarios compared to alternative methods in the literature.

We proved the validity of these advantages through empirical evaluations on synthetic datasets.
Moreover, we extensively evaluated both our original formulation and a fully-differentiable restriction when used as loss penalizers in fair ML contexts.
Experimental results obtained across three benchmark datasets confirm our hypothesis that our kernel-based $\operatorname{HGR}$ indicator can effectively provide meaningful gradient information for the training of neural models.

\bibliographystyle{cas-model2-names}
\bibliography{bibliography}

\appendix

\section{Pearson's Correlation as Least Squares}
\label{app:pearson_lstsq}

Let us start from the following least-square problem over standardized variables:
\begin{equation}
    \argmin_r \frac{1}{n} \left\| \frac{a - \mu_a}{\sigma_a} \cdot r\ - \frac{b - \mu_b}{\sigma_b} \right\|_2^2
\end{equation}
with $\mu_a, \mu_b$ and $\sigma_a, \sigma_b$ being the mean and standard deviations of vectors $a$ and $b$, respectively.
We know that the problem is convex as it simply features a vector product between the inputs and the variable $r$.
This means that the optimal solution can be achieved by setting the gradient with respect to $r$ to zero, i.e.: 
\begin{equation}
    \frac{1}{n} \left(\frac{a - \mu_a}{\sigma_a} \cdot r - \frac{b - \mu_b}{\sigma_b}\right)^T \frac{a - \mu_a}{\sigma_a} = 0
\end{equation}
which, after algebraic manipulation, can be rewritten as:
\begin{equation}
    \frac{1}{n} \frac{(a - \mu_a)^T (a - \mu_a)}{\sigma_a^2} \cdot r = \frac{1}{n} \frac{(a - \mu_a)^T (b - \mu_b)}{\sigma_a \cdot \sigma_b}
\end{equation}

With further observations, we can notice that the term $\frac{1}{n}(a - \mu_a)^T (a - \mu_a)$ denotes the variance $\sigma_a^2$.
We can therefore simplify the left side, thus arriving at:
\begin{equation}
    r = \frac{1}{n} \frac{(a - \mu_a)^T (b - \mu_b)}{\sigma_a \cdot \sigma_b}
\end{equation}
which corresponds in fact to the sample Pearson correlation coefficient.

\section{Simplification to a Single-Level Problem}
\label{app:hgr_single_level}

Consider the definition of HGR given in \Cref{eq:hgr_bilevel}.
Since HGR is based on Pearson's correlation, which is scale-independent, we can fix zero mean and unitary standard deviation without loss of generality, hence obtaining:
\begin{equation}
    \max_{f, g} \argmin_r \frac{1}{n} \left\| r \cdot f_a - g_b \right\|_2^2
    \qquad \text{s.t.} \qquad
    \mathbb{E}\left[f_a\right] = \mathbb{E}\left[g_b\right] = 0, \qquad
    \mathbb{E}\left[f^2_a\right] = \mathbb{E}\left[g^2_b\right] = 1
\end{equation}
where $f_a = f(a)$ and $g_b = g(b)$ are used as aliases to improve clarity.

We introduce two additional copula transformations $p_a = p(a)$ and $q_b = q(b)$, along with their related correlation coefficient $w$.
Assume that one transformation pair results in a lower Mean Squared Error, i.e.:
\begin{equation}
    \frac{1}{n} \left\| r \cdot f_a - g_b \right\|_2^2 < \frac{1}{n} \left\| w \cdot p_a - q_b \right\|_2^2
\end{equation}
we can further expand these terms as follows:
\begin{equation}
    \frac{f^T_a f_a}{n} r^2 - 2 \frac{f^T_a g_b}{n} r + \frac{g^T_b g_b}{n}
    <
    \frac{p^T_a p_a}{n} w^2 - 2 \frac{p^T_a q_b}{n} w + \frac{q^T_b q_b}{n}
\end{equation}

Given our zero-mean assumption, all quadratic terms such as $f^T_a f_a / n$ represent sample variances $\mathbb{E} \left[ f^2_a \right]$.
Consequently, since under the same assumptions variances are unitary, we can simplify this inequality as: \begin{equation}
    r^2 - 2 \frac{f^T_a g_b}{n} r + 1
    <
    w^2 - 2 \frac{p^T_a q_b}{n} w + 1
\end{equation}

We can further reduce this inequality by noting that $f^T_a g_b / n$ and $w = p^T_a q_b / n$ represent the correlation coefficients $r$ and $w$, respectively. We obtain: \begin{equation}
    r^2 - 2 r^2 + 1 < w^2 - 2 w^2 + 1
\end{equation}
which leads to:
\begin{equation}
    r^2 > w^2
\end{equation}

Given that all transformations are invertible, we can conclude that:
\begin{equation}
    \label{eq:hgr_equiv}
    r^2 > w^2
    \quad \Leftrightarrow \quad
    \frac{1}{n} \left\| r \cdot f_a - g_b \right\|_2^2 < \frac{1}{n} \left\| w \cdot p_a - q_b \right\|_2^2
\end{equation}
In essence, maximizing the square of the sample HGR equates to minimizing the Mean Squared Error. To maximize $r^2$, one needs to either maximize $r$ or minimize $-r$.
Given the flexible nature of copula transformations, the sign of $r$ can be altered by changing the sign of either $f$ or $g$.
Thus, maximizing $r$ is ultimately equivalent to maximizing $r^2$ in this context.

\section{Soundness of Lagrangian Formulation}
\label{app:lagrangian_soundness}

Let us consider a lagrangian formulation where $c(\cdot)$ represents the objective function and $p(\cdot)$ is the penalty function.
Given two distinct multipliers, $\mu$ and $\nu$, we obtain two separate solutions to the corresponding minimization problems:
\begin{align}
    m & \in \argmin_x \{ c(x) + \mu \cdot p(x) \} \\
    n & \in \argmin_x \{ c(x) + \nu \cdot p(x) \}
\end{align}

As both $m$ and $n$ are optimal for their respective problems, it follows that:
\begin{align}
    c(m) + \mu \cdot p(m) \leq c(n) + \mu \cdot p(n) \quad & \implies \quad \mu \cdot [p(m) - p(n)] \leq c(n) - c(m) \\
    c(m) + \nu \cdot p(m) \geq c(n) + \nu \cdot p(n) \quad & \implies \quad \nu \cdot [p(m) - p(n)] \geq c(n) - c(m)
\end{align}

Then, by combining the previous two equations, we obtain:
\begin{equation}
    \label{eq:lagrangian_relationship}
    \mu \cdot [p(m) - p(n)] \leq c(n) - c(m) \leq \nu \cdot [p(m) - p(n)]
\end{equation}

Let us now recall our problem described in \Cref{eq:hgr_constrained}.
The objective function is formulated as:
\begin{equation}
    \begin{split}
        c(\alpha, \beta) & = \left\| \widetilde{\mathbf{P}}_a^h \cdot \widetilde{\alpha} - \widetilde{\mathbf{P}}_b^k \cdot \beta \right\|_2^2 = \\
                         & = (\widetilde{\mathbf{P}}_a^h \cdot \widetilde{\alpha} - \widetilde{\mathbf{P}}_b^k \cdot \beta)^T \cdot
                             (\widetilde{\mathbf{P}}_a^h \cdot \widetilde{\alpha} - \widetilde{\mathbf{P}}_b^k \cdot \beta) = \\
                         & = (\widetilde{\mathbf{P}}_a^h \cdot \widetilde{\alpha})^T \cdot (\widetilde{\mathbf{P}}_a^h \cdot \widetilde{\alpha}) - 
                             2 \cdot (\widetilde{\mathbf{P}}_a^h \cdot \widetilde{\alpha})^T \cdot (\widetilde{\mathbf{P}}_b^k \cdot \beta) +
                             (\widetilde{\mathbf{P}}_b^k \cdot \beta)^T \cdot (\widetilde{\mathbf{P}}_b^k \cdot \beta) = \\
                         & = \sum_{i=1}^h \sum_{i=1}^h \alpha_i \alpha_j \cdot \operatorname{cov}(a^i, a^j) -
                             2 \cdot \sum_{i=1}^h \sum_{i=1}^k \alpha_i \beta_j \cdot \operatorname{cov}(a^i, b^j) +
                             \sum_{i=1}^k \sum_{i=1}^k \beta_i \beta_j \cdot \operatorname{cov}(b^i, b^j)
    \end{split}
\end{equation}

Similarly, we can build a convex penalty function which evaluates to zero when the constraint $\sigma(\mathbf{P}_b^k \cdot \beta) = 1$ is satisfied:
\begin{equation}
    \begin{split}
        p(\alpha, \beta) & = \left| \sigma(\mathbf{P}_b^k \cdot \beta)^2 - 1 \right| = \\
                         & = \left| \sum_{i=1}^k \sum_{i=1}^k \beta_i \beta_j \cdot \operatorname{cov}(b^i, b^j) - 1 \right|
    \end{split}
\end{equation}

Given that we are dealing with data samples, we can assume input vectors $a$ and $b$ to have strictly finite variance.
Therefore, both $c(\cdot)$ and $p(\cdot)$ can evaluate to infinity if and only if at least one coefficient $\alpha_i$ or $\beta_i$ is infinite, i.e.:
\begin{align}
    c(\alpha, \beta) \rightarrow \pm \infty \iff \exists \alpha_i = \pm \infty \vee \exists \beta_i = \pm \infty \\
    p(\alpha, \beta) \rightarrow + \infty \iff \exists \alpha_i = \pm \infty \vee \exists \beta_i = \pm \infty
\end{align}

Nonetheless, we are assuming $\forall \alpha_i, \beta_i \in \mathbb{R}$, hence neither of the two functions can take infinite values.
Going back to \Cref{eq:lagrangian_relationship}, we can conclude that, for any $p(m), p(n), \mu, \nu \in \mathbb{R}$:
\begin{equation}
    \label{eq:lagrangian_conclusion}
    \mu \cdot [p(m) - p(n)] \leq \nu \cdot [p(m) - p(n)] \quad \implies \begin{cases}
        p(m) > p(n) \implies p(m) - p(n) > 0 \implies \nu \geq \mu \\
        p(m) < p(n) \implies p(m) - p(n) < 0 \implies \nu \leq \mu
    \end{cases}
\end{equation}
or rather that, if a more constrained solution exists, it could be obtained using a strictly finite multiplier $\nu \geq \mu \in \mathbb{R}$.
As a consequence, the considered problem admits a penalty formulation that: 1) is based on convex subproblems; 2) asymptotically converges to an optimal solution as the associated multiplier grows.

\section{Monotonicity of HGR-KB}
\label{app:proof_monotonicity}

Let $\widetilde{\alpha}$ and $\beta$ represent the optimal coefficients derived from the computation of $\operatorname{HGR-KB}(a, b; h, k)$.

Given $h^\prime \geq h$ and $k^\prime \geq k$, it follows that:
\begin{equation}
    \operatorname{HGR-KB}(a, b; h^\prime, k^\prime) < \operatorname{HGR-KB}(a, b; h, k) \iff \nexists \widetilde{\alpha}^\prime, \beta^\prime \text{ s.t. } \rho(\mathbf{P}_a^{h^\prime} \cdot \widetilde{\alpha}^\prime, \mathbf{P}_b^{k^\prime} \cdot \beta^\prime) \geq \rho(\mathbf{P}_a^h \cdot \widetilde{\alpha}, \mathbf{P}_b^k \cdot \beta)
\end{equation}

However, this assertion is invalid, as we there exist vectors $\widetilde{\alpha}^\prime = \begin{pmatrix} \widetilde{\alpha} & \mathbf{0}_{h^\prime - h}\end{pmatrix}$ and $\beta^\prime = \begin{pmatrix} \beta & \mathbf{0}_{k^\prime - k}\end{pmatrix}$ which, by nullifying the influence of higher degrees, results in the same value of $\operatorname{HGR-KB}(a, b; h, k)$.
This property is depicted in \Cref{fig:monotonicity}, where we empirically validate it across the three benchmarks used in our experiments.

\end{document}